%% file: main.tex
\documentclass[10pt,twocolumn,letterpaper]{article}

\usepackage{3dv}
\usepackage{times}
\usepackage{epsfig}
\usepackage{graphicx}
\usepackage{amsmath}
\usepackage{amssymb}
\usepackage{caption,subcaption}
\usepackage[super]{nth}
\usepackage{siunitx}
\usepackage{booktabs}
\usepackage{etoolbox}
\usepackage{enumitem}
\usepackage{multirow}
\usepackage{verbatim}
\usepackage{xcolor}
\usepackage{array}
\usepackage[normalem]{ulem}
\usepackage[accsupp]{axessibility}  

\usepackage[pagebackref=true,breaklinks=true,colorlinks,bookmarks=false]{hyperref}

\threedvfinalcopy 

\def\threedvPaperID{106} 

\ifthreedvfinal\fi

\setlist{noitemsep}

\newcommand{\PAR}[1]{\vskip4pt \noindent{\bf #1~}}

\renewrobustcmd{\bfseries}{\fontseries{b}\selectfont}
\renewrobustcmd{\boldmath}{}
\newrobustcmd{\B}{\bfseries}

\newcommand{\ra}[1]{\renewcommand{\arraystretch}{#1}}
\newcolumntype{H}{>{\setbox0=\hbox\bgroup}c<{\egroup}@{}}

\usepackage[font={small}]{caption}

\begin{document}

\title{
\vspace{-1em}
LatentHuman: Shape-and-Pose Disentangled Latent Representation \\ for Human Bodies
\vspace{-1em}
}

\author{Sandro Lombardi$^{1*}$ \quad Bangbang Yang$^{2*}$ \quad Tianxing Fan$^{2}$ \\
Hujun Bao$^{2}$ \quad Guofeng Zhang$^{2}$ \quad Marc Pollefeys$^{1,3}$ \quad Zhaopeng Cui$^{2\dag}$ \and
$^{1}$ETH Zurich \quad
$^{2}$State Key Lab of CAD\&CG, Zhejiang University  \quad
$^{3}$Microsoft
}

\twocolumn[{%
\maketitle
\begin{center}
    \addtolength{\belowcaptionskip}{0.5em}
    \centering
    \vspace{-2.0em}
    \captionsetup{type=figure}
    \includegraphics[width=0.92\textwidth]{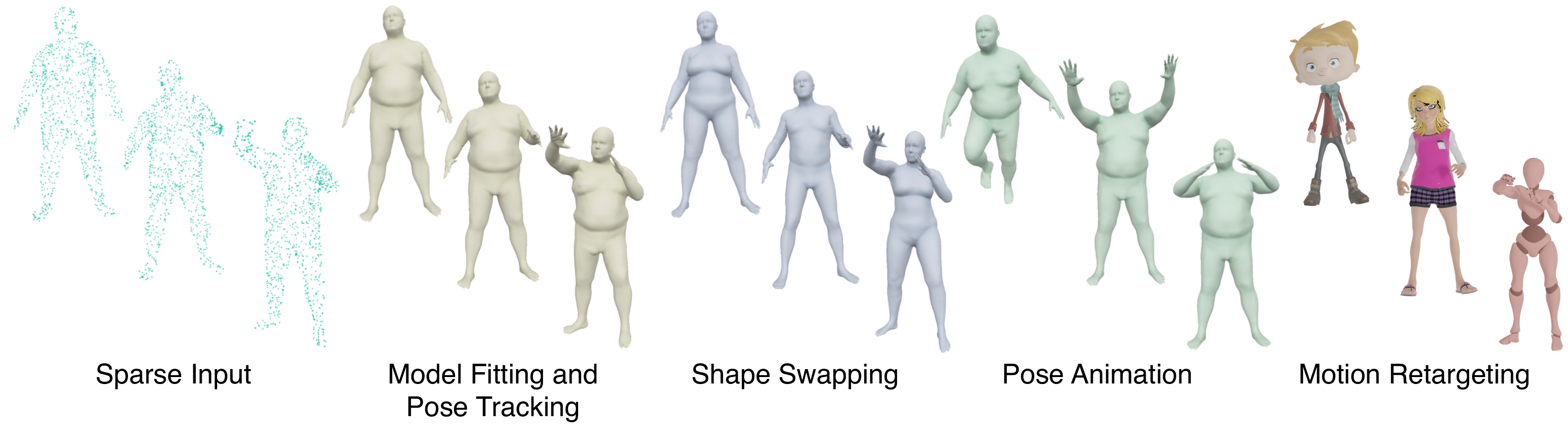}
    \captionof{figure}{
    \textbf{LatentHuman.} 
    We propose a novel representation for the human body which leverages piecewise neural implicit functions to learn shape and pose disentangled latent embeddings.
    With the fully differentiable and optimizable design as well as the separated latent spaces, LatentHuman can be applied to several 3D learning tasks, including 3D model fitting  (model at the top row in second column) and pose tracking (models at the bottom two rows in second column) from sparse input point clouds, shape swapping, pose animation, and motion retargeting. 
    }
    \label{fig:teaser}
\end{center}%
 }]

{
  \renewcommand{\thefootnote}%
    {\fnsymbol{footnote}}
  \footnotetext{$^{\dag}$ Corresponding author}
  \footnotetext{$^{*}$ Equal contributions}
  \footnotetext{$\;\;$ Project webpage: \url{https://latenthuman.github.io/}}
}


\thispagestyle{empty}

\begin{abstract}
3D representation and reconstruction of human bodies have been studied for a long time in computer vision.
Traditional methods rely mostly on parametric statistical linear models, limiting the space of possible bodies to linear combinations.
It is only recently that some approaches try to leverage neural implicit representations for human body modeling, and while demonstrating impressive results, they are either limited by representation capability or not physically meaningful and controllable.
In this work, we propose a novel neural implicit representation for the human body, which is fully differentiable and optimizable with disentangled shape and pose latent spaces.
Contrary to prior work, our representation is designed based on the kinematic model, which makes the representation controllable for tasks like pose animation, while simultaneously allowing the optimization of shape and pose for tasks like 3D fitting and pose tracking. 
Our model can be trained and fine-tuned directly on non-watertight raw data with well-designed losses. Experiments demonstrate the improved 3D reconstruction performance over SoTA approaches and show the applicability of our method to shape interpolation, model fitting, pose tracking, and motion retargeting.
\end{abstract}
\vspace{-2.0em}


\section{Introduction}
Modeling of 3D human bodies has been studied extensively. The mesh representation is adopted in most of existing methods including traditional parametric models 
with PCA decomposed shape space and kinematic tree-based pose space
~\cite{loper_smpl_2015,osman_star_2020,pavlakos_expressive_2019,romero_embodied_2017} 
as well as more sophisticated learning-based models with deformed mesh-based templates~\cite{tretschk_demea_2020,jiang_disentangled_2020,xu_ghum_2020}. However, the mesh representation is normally limited by the fixed topology.

Recently, the articulated neural implicit representation~\cite{deng_nasa_2020,mihajlovic_leap_2021} has been proposed for human bodies. However, these methods either model human poses for a single human shape~\cite{deng_nasa_2020}, or rely on the SMPL~\cite{loper_smpl_2015} model shape parameters~\cite{mihajlovic_leap_2021}, which limits the representation capability of these models. 
As a concurrent work, Palafox~\etal~\cite{palafox_npms_2021} propose to represent the shape and the pose of human bodies with disentangled latent codes, which enable optimization to fit new observations at test time.
Nevertheless, their latent pose code encodes a deformation field rather than a physically meaningful human pose and thus is not suitable for controlling human bodies.
Moreover, all these methods require watertight 3D models for training which is difficult to generate for real-world scans.

In this paper, we introduce LatentHuman, a neural representation for human bodies with disentangled meaningful pose and shape representations like traditional parametric models, which is fully differentiable and facilitates a variety of applications, \eg, 3D model fitting, 3D pose tracking, shape swapping, pose animation, and motion retargeting, etc (see Fig.~\ref{fig:teaser}).
LatentHuman fully leverages the power of neural implicit functions and represents pose and shape as differentiable and optimizable latent codes. 
First, in contrast to the methods using one global latent code for the whole body~\cite{gropp_implicit_2020,atzmon_sal_2020a}, the disentanglement of pose and shape spaces allows us to optimize them separately. 
Furthermore, the representation is differentiable end-to-end and can be used in deep learning pipelines. 
This property, together with the embedding of the shape and pose as latent codes, enables tasks which require optimization - either for the shape, the pose, or both - which is critical, \eg, in 3D model fitting and pose tracking.
Additionally, we allow the pose to be parameterized by traditional pose parameters, \ie, used by SMPL~\cite{loper_smpl_2015}, to make the representation controllable, which is an important property for tasks like motion retargeting to cartoon characters.
Lastly, we make use of implicit geometric regularization \cite{gropp_implicit_2020} in order to train our models on raw input point clouds. This allows us to fine-tune our model on unprocessed raw-scan datasets with point-cloud or depth map input without requiring the watertightness property.

However, disentangling the shape and pose factors is hard because the shape of a human body usually deforms with different poses.
In order to fix this problem, we guide the learning of pose, shape properties and skeleton joints by exploiting the kinematic model which introduces a constraint on pose and skeleton.  
More specifically, the VPoser module~\cite{pavlakos_expressive_2019} is adopted to learn the latent pose representation, while a novel VJointer module is proposed for learning the latent skeleton joint representation as one part of the shape representation. 
By taking kinematic models into consideration, we can generate meaningful human poses and skeletons. 
In order to model detailed human shapes even with clothes, we utilize the piecewise deformable model proposed by Deng~\etal~\cite{deng_nasa_2020} to learn pose-dependent deformations and the other part of the shape representation responsible for shape identities.
A novel dual-weighting mechanism is proposed to guarantee smooth and gradual supervision for part connections, which results in a seamless body reconstruction.
Furthermore, we adopt similar loss definitions as proposed in Gropp~\etal~\cite{gropp_implicit_2020} and  SIREN~\cite{sitzmann_implicit_2020} to learn the disentangled shape and pose representation directly on non-watertight data.
Different from the unstructured model used in \cite{gropp_implicit_2020, sitzmann_implicit_2020}, our piecewise deformable model is more complex and harder to train since one body part model may also influence other parts. 
In order to prevent the individual human body parts from growing into neighbouring parts, we propose a novel one-sided loss that exploits the nature of body part placement and guides the piecewise functions to focus on their own body part.

We demonstrate the viability of LatentHuman on 3D learning tasks, including human body representation, shape interpolation, 3D model fitting, 3D pose tracking, motion retargeting, and fine-tuning on partial raw scans. With the learnt latent spaces for shape and pose, LatentHuman is able to effectively produce valid shape predictions or pose configurations for interpolation while being readily optimizable for model fitting and pose tracking.

The main contributions of this paper can be summarized as follows:
\begin{itemize}[itemsep=1pt,topsep=1pt,leftmargin=*]
\item We propose LatentHuman, a novel shape-pose-disentangled representation for human bodies. 
Our representation uses a neural implicit function per body part and encapsulates pose and shape in latent embeddings based on the kinematic model, so the pose is controllable through SMPL-based pose parameters.
\item Our representation is fully differentiable end-to-end and can be optimized during inference in deep learning pipelines. 
\item We propose a novel dual-weighting mechanism to guarantee a smooth piecewise deformable model, and a novel one-sided loss to enable training directly on non-watertight data.
\item Extensive experiments show that LatentHuman has better representation ability and robustness compared to existing methods.

\end{itemize}

\section{Related work}
\PAR{Neural Implicit Representations.} 
Recent neural implicit representations, \ie, deep coordinate-based MLP networks, have been proposed for 3D modeling as a light-weight alternative to traditional 3D representations like surfel~based~\cite{schops_surfelmeshing_2019,whelan_elasticfusion_2016}, mesh-based~\cite{bagautdinov_modeling_2018,baque_geodesic_2018} and voxel-based methods~\cite{curless_volumetric_1996,newcombe_kinectfusion_2011,fuhrmann_fusion_2011}.
While original pioneering works were limited to small-scale objects with a lack of details~\cite{park_deepsdf_2019,mescheder_occupancy_2019,chen_learning_2018,michalkiewicz_deep_2019}, follow-up works have addressed the issues with self-supervision~\cite{atzmon_sal_2020,atzmon_sal_2020a,gropp_implicit_2020}, open surfaces~\cite{chibane_neural_2020}, room-scale reconstruction~\cite{chabra_deep_2020,peng_convolutional_2020,jiang_local_2020,lombardi_scalable_2020,mi_ssrnet_2020}, training procedure~\cite{duan_curriculum_2020,sitzmann_metasdf_2020} or high-frequency details~\cite{sitzmann_implicit_2020}.

\PAR{Classical Parametric Models.}
In contrast to generic 3D shapes, articulated shapes have traditionally been learned from registered meshes, accomplished with a skinning algorithm that deforms vertices of a mesh surface as the joints of an underlying skeleton change~\cite{jacobson_skinning_2014,wang_multiweight_2002,kavan_spherical_2005}, outperforming the straight-forward and widely used linear blend skinning (LBS) algorithm.
Loper \etal~\cite{loper_smpl_2015} then improved significantly upon prior works by introducing a skinned multi-person linear model (SMPL). 
While SMPL only models the human body, related or follow-up works have focused on other shape parts like faces~\cite{blanz_morphable_1999,dai_3d_2017,li_learning_2017,paysan_3d_2009,ploumpis_combining_2019} and hands~\cite{romero_embodied_2017}, a combination of those~\cite{joo_total_2018,pavlakos_expressive_2019,romero_embodied_2017},  animals~\cite{zuffi_3d_2017} or improving on issues, \eg, long-distance inter-dependencies across the skeleton~\cite{osman_star_2020}.

\PAR{Learnt Mesh-based Articulated Representations.}
Recent works have started to adopt deep learning-based approaches for modeling articulated shapes.
Some works focus on using template meshes with auto-encoder like structures~\cite{xu_ghum_2020,tretschk_demea_2020, zhou_unsupervised_2020,jiang_disentangled_2020}. 
Such methods have the advantage of always predicting topologically correct outputs but are restricted to the predefined topology of the template. 
Other works based on NeRF~\cite{mildenhall_nerf_2020} rely on meshes for novel view synthesis~\cite{peng_neural_2021} or rendering of avatars~\cite{prokudin_smplpix_2020}, although the work from Peng \etal~\cite{peng_neural_2021} can be seen as a hybrid: On one hand, using a mesh for anchoring of latent features and on the hand using those features as input for neural implicit functions.
Our method focuses on human shape representation from a geometric perspective as we specifically model various human shapes and enable shape-related tasks such as 3D model fitting, pose tracking and motion retargeting.

\PAR{Articulated Neural Implicit Representations.}
While some early neural implicit representation methods demonstrated compelling results on human body scans~\cite{atzmon_sal_2020a,gropp_implicit_2020,chibane_implicit_2020}, Deng~\etal~\cite{deng_nasa_2020} pointed out that a straightforward application of auto-decoders for encoding posed shapes into one global latent code~\cite{gropp_implicit_2020, atzmon_sal_2020a} is often not enough to cope with the large variety of possible human body poses. Instead they propose NASA, a piecewise deformable model which naturally handles large deformations of body poses. 
They use joint transformation matrices of a kinematic tree to warp input query points prior to running them through a neural implicit function network. 
Notably, they use one network per joint to learn pose-dependent, canonical body part shapes.

Subsequent works can be distinguished by the the way they warp between a canonical shape space and a posed shape space.
The concurrent work SNARF~\cite{chen2021snarf}, which learns a forward skinning weight field, improves upon NASA, especially for out-of-distribution examples.
They find a many-to-one mapping of possible canonical point candidates for a given point in the deformed space through iterative optimization of the implicit forward skinning equation.
However, for both methods, their model is conditioned on poses only and tied to the same shape identity, needing retraining for new subjects.
Different from NASA and SNARF, we aim to represent human bodies of arbitrary poses and shapes.
SCANimate~\cite{saito_scanimate_2021} omits per-joint implicit functions and rather trains one big neural implicit decoder for representing the shape in canonical space. In addition, they learn forward and inverse skinning networks with cycle-consistency constraints which they use for posing the shape. 
The idea of learning skinning weights was also simultaneously suggested by Mihajlovic \etal~\cite{mihajlovic_leap_2021}. 
They propose LEAP, a method for learning the articulated occupancy of human bodies. They build upon SMPL~\cite{loper_smpl_2015} and provide efficient occupancy checks on query points. Query points are mapped to the canonical space through a trained inverse LBS network which learns valid skinning weights for every query point, even non-manifold points.
However, they heavily rely on SMPL shape parameters which restricts the representation capability. In contrast, our method learns the shape latent space from scratch.

The concurrent work NPMs~\cite{palafox_npms_2021} also learns to disentangle shape and pose in a fully implicit manner with several latent codes.
Their shape latent space resides in the canonical space while their learned deformation field maps to the posed space.
In contrast to NPMs, our representation additionally allows the use of well-known SMPL pose parameters to control the pose latent code which is useful for animation tasks. 
Further, while NPMs need supplementary initialization networks and depth-projected SDF grids to infer and optimize the pose and shape latent codes, we instead rely on implicit geometric regularization~\cite{gropp_implicit_2020} allowing us to work with raw, non-watertight point clouds at training time.
A summary of recent related works is shown in Table~\ref{tab:contribution}.
\input{tables/tab_contributions}

\section{Method}
\label{sec:method}
\newcommand{\Xparam}{\ensuremath{\theta}} 
\newcommand{\XS}{\ensuremath{S}} 
\newcommand{\XP}{\ensuremath{P}} 
\newcommand{\XN}{\ensuremath{N}} 
\newcommand{\XX}{\ensuremath{X}} 
\newcommand{\Xfft}{\ensuremath{\gamma}} 
\newcommand{\Xl}{\ensuremath{l}} 
\newcommand{\XK}{\ensuremath{K}} 

\newcommand{\Xspred}{\ensuremath{\tilde{\Xs}}} 
\newcommand{\Xstd}{\ensuremath{\sigma}} 
\newcommand{\Xvar}{\ensuremath{\Xstd^{2}}} 
\newcommand{\Xthres}{\ensuremath{\delta}} 
\newcommand{\abs}[1]{\left|#1\right|}
\newcommand{\norm}[1]{\left\lVert#1\right\rVert}
\newcommand{\clamp}{\ensuremath{\text{clamp}}} 
\newcommand{\R}{\mathbb{R}}
\newcommand{\XsmplPose}{\ensuremath{\theta}} 
\newcommand{\Xz}{\ensuremath{\mathbf{z}}} 
\newcommand{\XM}{\ensuremath{M}} 
\newcommand{\XQ}{\ensuremath{Q}} 
\newcommand{\Xx}{\ensuremath{\mathbf{x}}} 
\newcommand{\Xn}{\ensuremath{\mathbf{n}}} 
\newcommand{\XSDF}{\ensuremath{\text{SDF}}} 
\newcommand{\XDomain}{\ensuremath{\Omega}} 
\newcommand{\Xnet}{\ensuremath{\XSDF}} 
\newcommand{\Xloss}{\ensuremath{\mathcal{L}}} 
\newcommand{\Xs}{\ensuremath{s}} 
\newcommand{\XB}{\ensuremath{B}} 
\newcommand{\Xb}{\ensuremath{b}} 
\newcommand{\XBoneT}{\ensuremath{\mathbf{B}}} 
\newcommand{\Xrootj}{\ensuremath{\mathbf{t}_{0}}} 
\newcommand{\XprojLayer}{\ensuremath{\Pi}} 

\begin{figure*}[!htbp]
    \addtolength{\belowcaptionskip}{-1.0em}
    \centering
    \includegraphics[width=0.85\linewidth, trim={0 0 0 0}, clip]{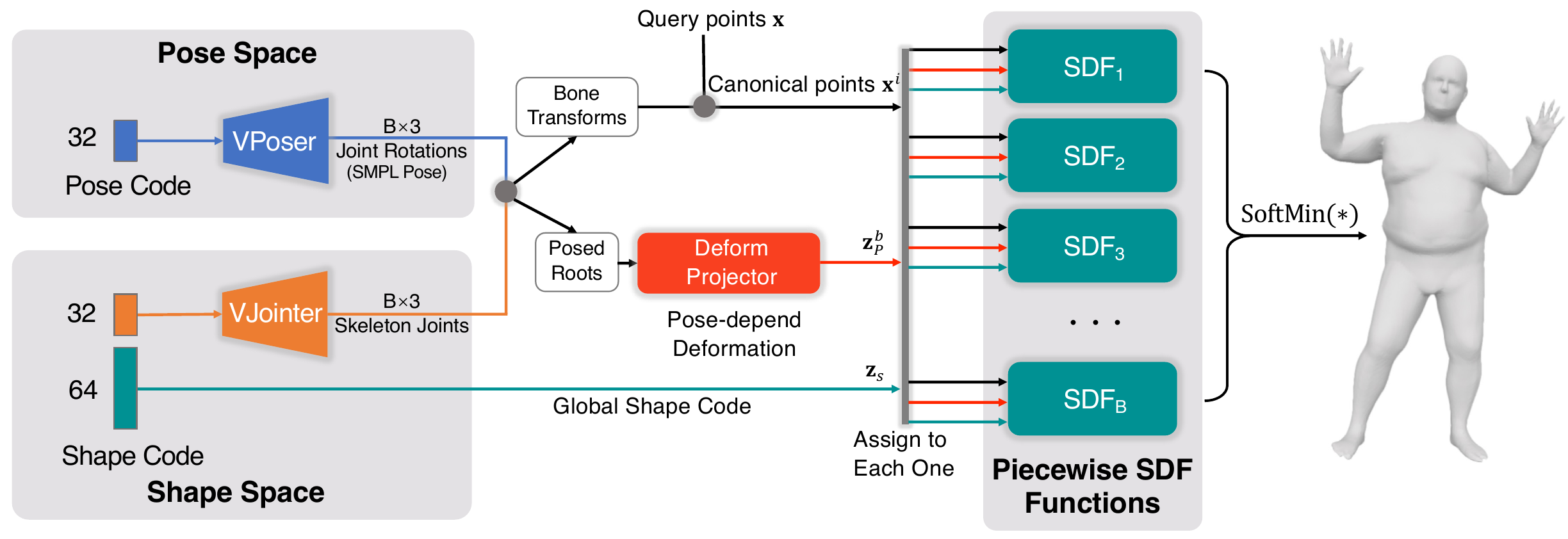}
    \caption{
    \textbf{Overview.} 
    For learning SDF functions per body part, we use a piecewise deformable model~\cite{deng_nasa_2020} conditioned on a shape code $\Xz_{s}$, a per-joint feature $\Xz_{p}^{b}$ describing pose-dependent deformations and canonical query points $\Xx^{i}$.
    The bone transformations needed for obtaining $\Xz_{p}^{b}$ can be computed with SMPL pose joint rotations and skeleton joints in canonical space (\ie T-pose). For the former, we adopt VPoser~\cite{pavlakos_expressive_2019} while for the latter, we introduce our novel VJointer module.
    The per-joint SDF predictions are combined with a SoftMin function to obtain the SDF values of the final mesh.
    }
    \label{fig:framework}
\end{figure*}

We introduce LatentHuman, a fully differentiable and disentangled implicit representation for human bodies.
Following prior works~\cite{park_deepsdf_2019,atzmon_sal_2020,gropp_implicit_2020,sitzmann_implicit_2020}, we express the body shape through signed distance functions (SDF) with feed-forward neural networks.
As shown in Fig.~\ref{fig:framework}, we consider two spaces which influence the posed human body model and represent both with latent codes respectively:
The pose code controls the joint movement of the body while the shape code determines the canonical skeleton joints and specific body shapes (Sec~\ref{ssec:meth_disentangled_pose_shape}).
For a given query point $\Xx_{i}$, LatentHuman predicts per-joint canonical SDF values with a series of piecewise implicit functions,
which are conditioned on shape and pose-dependent deformation features (Sec~\ref{ssec:meth_implicit_representation}).
Motivated by Gropp~\etal~\cite{gropp_implicit_2020}, we adopt implicit geometric regularization and related auxiliary loss functions to enable training and fine-tuning on non-watertight raw scans. To make the training procedure stable and generate seamless body shapes, we introduce a novel one-sided non-manifold loss and a dual-weighting strategy (Sec~\ref{ssec:non_rigid_geometric_supervision}).

\subsection{Disentanglement of Pose and Shape}
\label{ssec:meth_disentangled_pose_shape}
As the human shape normally deforms with different poses, we exploit the kinematic model and prior knowledge to learn the disentanglement of human pose and shape.

\PAR{Pose Space.}
Ideally, a human pose encoding should  allow valid poses only, be human-readable and reusable in downstream tasks. Latent embeddings~\cite{park_deepsdf_2019,palafox_npms_2021},
in contrast to kinematic tree skeletons, are usually not human-readable and not friendly for downstream tasks like motion animation.
To this end, we adopt the kinematic pose prior VPoser~\cite{pavlakos_expressive_2019} as our pose space decoder and use the joint rotations (\ie, $\XsmplPose \in \R^{B\times3}$, where $B$ is the number of body parts) of the kinematic tree~\cite{loper_smpl_2015, hirshberg_coregistratio_2012}  as output in order to compute bone transformations used by our piecewise deformable model.
This allows us to either use SMPL pose parameters for rigging human bodies or optimizing the pose in latent space, which yields a more robust performance on downstream tasks like pose tracking.

\PAR{Shape Space.}
Unlike the human pose space, the shape space needs to store more information since human shapes can be more expressive especially when dealing with clothed bodies.
Meanwhile, it is known that the human shape structure, including overall height, length of body parts, and relative positioning of body parts with respect to each other, follows the kinematic model and is independent of surface-related details. If a single latent code is used to encode the whole shape, the human shape structure will be deeply intertwined with other shape properties, \eg, the human's body weight index or other surface-related properties like wrinkles, and thus we cannot learn it separately.

For this reason, we split the shape encoding into two components, one for representing the joints of the skeleton in canonical space, \ie, the T-pose,
and one for representing the remaining shape properties.
We assume that the distribution of canonical skeleton joints lies on a lower-dimensional subspace and therefore, inspired by VPoser~\cite{pavlakos_expressive_2019}, we propose a data-driven kinematic prior named VJointer, which learns a human joint prior with a variational autoencoder (VAE)~\cite{kingma_autoencoding_2013} on the SURREAL dataset~\cite{surreal}.
By introducing  VJointer, we can benefit from regularized optimization of skeleton joints in a 32-dimensional latent space.
For the surface-related shape features, we use a 64-dimensional learnable shape code $\Xz_{s}$ to encode different human body identities.

\subsection{Implicit Human Body Representation}
\label{ssec:meth_implicit_representation}
Inspired by NASA~\cite{deng_nasa_2020}, we use a series of piecewise implicit functions to encode the human body. 
Different from NASA that adopts occupancy functions,
we represent human bodies with implicit SDF functions which encode the distance to the closest body surface for a 3D query point $\Xx$, with the sign indicating if the point is inside (negative) or outside (positive) the body.
Each SDF function outputs a body part SDF value conditioned on a pose-dependent deformation feature $\Xz_{p}^{b}$ and the corresponding shape code $\Xz_{s}$.
All predicted part SDF values are finally blended to obtain the overall SDF value for the whole body.

\PAR{Pose-dependent Deformation.}
To encode the pose-dependent deformation feature $\Xz_{p}^{b}$, we follow NASA~\cite{deng_nasa_2020} and transform the root bone location $\Xrootj$ to the per-bone local spaces and project to a part-specific feature vector with a learnable linear layer $\XprojLayer_{b}: \R^{\XB\times3}\rightarrow\R^{32}$:
\begin{equation}
\Xz_{p}^{b}=\XprojLayer_{b}\left(\oplus_{b}^{B} \XBoneT_{b}^{-1}\Xrootj\right),
\end{equation}
where $\oplus$ concatenates each transformed root.

\PAR{Piecewise Model.}
We consider $B$ body parts and for each employ a small coordinate-based MLP conditioned on the pose-dependent deformation code $\Xz_{p}^{b}$ and the shape latent code $\Xz_{s}$. 
We transform query points $\Xx_i$ to the canonical T-pose according to the per-bone transformations $\XBoneT_{b}^{-1}$ and use those canonical query coordinates $\hat{\Xx}_i=\XBoneT_{b}^{-1}\Xx_i$ as input for each part MLP.
To further improve high-frequency surface details, we also apply a positional encoding $\gamma(\cdot)$~\cite{tancik_fourier_2020} to the canonical input coordinates and adopt a progressive training scheme~\cite{nerfies} to ensure a stable convergence.
For each implicit function, we output SDF values for the respective body part $b \in \{1,...,B\}$:
\begin{equation}
\label{eq:sdf}
  \hat{s}_{b}^{i} = \XSDF_{b}(\gamma(\hat{\Xx}_{i}) | \Xz_{p}^{b} , \Xz_{s}).
\end{equation}
The overall SDF value $\hat{s}_{i}$ can then be obtained as:
\begin{equation}
\label{eq:blend}
\hat{s}_{i} = \text{SoftMin}_{b}^{B}(\lambda_b \hat{s}_{b}^{i}),
\end{equation}
where $\lambda_b$ controls the blending weight of each body part.
We empirically set $\lambda_b = 50$, but a larger $\lambda_b$ also works well in our experiments.

\subsection{Non-rigid Geometric Supervision}
\label{ssec:non_rigid_geometric_supervision}
To make our representation trainable on non-watertight raw scans, we employ a series of supervision strategies.

\PAR{Supervision Sampling.}
For each ground-truth body mesh,
we sample a set of body part-weighted query points $\{\Xx_{i}\}^{2\XM+\XQ}$ consisting of $\XM$ points on the surface, $\XM$ near-surface points and $\XQ$ random points sampled within the bounding boxes of each body part. For our experiments, we use $\XM=4000$ and $\XQ=800$. 
As our training dataset consists of organized SMPL models, we use the provided skinning weights to assign to each query point the part label with the highest skinning weight.
Please refer to the supplementary material for more details.

\PAR{Geometric Supervision.}
Recent works~\cite{deng_nasa_2020, mihajlovic_leap_2021} mainly use densely supervised data to regress occupancy values, which requires a watertight mesh to ensure a successful ground-truth occupancy sampling.
To achieve the supervision directly on partial raw scan data,
we employ implicit geometric regularization as introduced by Gropp \etal~\cite{gropp_implicit_2020} and Sitzmann \etal~\cite{sitzmann_implicit_2020} by minimizing the following loss:
\begin{align}
\label{eq:geom_losses}
\Xloss_{\text{geom}} &= \lambda_{\text{m}}\Xloss_{\text{m}} + \lambda_{\text{n}}\Xloss_{\text{n}} + \lambda_{\text{e}}\Xloss_{\text{e}} + \lambda_{\text{nm}}\Xloss_{\text{nm}},\\
\Xloss_{\text{m}}&=\sum_{\Xx_i \in \XDomain_{0}} \phi(\Xx_i),\quad \phi(\Xx_i)=\abs{\Xnet(\Xx_i)},\\
\Xloss_{\text{n}}&=\sum_{\Xx_i \in \XDomain_{0}}(1-\langle\nabla_{\Xx_i}\Xnet(\Xx_i),\Xn(\Xx_i)\rangle),\\
\Xloss_{\text{e}}&=\sum_{\Xx_i \in \XDomain}\abs{\norm{\nabla_{\Xx_i}\Xnet(\Xx_i)}_{2}-1},\\
\Xloss_{\text{\text{nm}}}&=\sum_{\Xx \in \XDomain\setminus\XDomain_{0}}\exp(-\alpha\cdot\abs{\Xnet(\Xx)}),
\end{align}
where the manifold loss $\Xloss_{\text{m}}$ enforces points on the surface to yield zero SDF value,
with $\XDomain_{0}$ denoting the zero-level set.
$\Xloss_{\text{n}}$ is the normal loss that constrains the on-surface points to have a consistent gradient with the ground-truth surface normals $\Xn(\Xx)$.
The eikonal loss $\Xloss_{\text{e}}$ regularizes the norm of the spatial gradients to 1, where $\XDomain$ denotes the whole domain of the sampling.
$\Xloss_{\text{\text{nm}}}$ is the non-manifold loss which penalizes close-to-zero SDF values at off-surface points, where $\alpha$ is a hyper parameter and we set $\alpha=5$.
In practice, we add the above losses to both the SDF output of each piecewise SDF function (Eq.~\ref{eq:sdf}) and the overall blended SDF values (Eq.~\ref{eq:blend}).

\PAR{One-sided Non-manifold Loss.}
The aforementioned geometric supervision works for rigid shape learning~\cite{gropp_implicit_2020, sitzmann_implicit_2020}, but we find it does not ensure good convergence in our piecewise non-rigid case.
This is due to the lack of supervision signals on overlapping intersection areas at part connections, \ie, the outputs of the piecewise functions are unconstrained, which might lead one body part to extrapolate surface into neighboring parts.
We therefore introduce a one-sided non-manifold loss, through which we can penalize the prediction of negative (inside) SDF values for on-surface sampled points not belonging to the current body part:
\begin{equation}
\Xloss_{\text{osnm}}=\sum_{\Xx_i \in \XDomain_{0}^{j\neq b}}\text{SoftPlus}_{\beta}\left(\frac{-\Xnet_{b}(\Xx_i)+\Xthres}{2\Xthres}\right),
\end{equation}
where $\Xthres=0.01$ denotes a truncation distance and $\beta=10$ is the SoftPlus parameter.
We use SoftPlus rather than ReLU in order to allow smooth gradient flows. 

\PAR{Dual-weighting.}
Although the aforementioned part supervision enforces each SDF function to focus on their own body parts,
it will inevitably lead to joint artifacts near the part connection areas (see Fig.~\ref{fig:dual_weighting}).
Therefore, we introduce a dual-weighting mechanism by exploiting the skinning weights of the SMPL model to eliminate those joint artifacts and facilitate seamless body reconstruction.
More specifically, for each surface sampling point $\Xx_{i}$, we find two body part labels with the highest skinning weights $w^{0}_i$ and $w^{1}_i$.
We can then rewrite the part manifold loss in a dual-weighting manner:
\begin{equation}
\Xloss_{\text{dual-m}}=\sum_{\Xx \in \XDomain_{0}^{j\neq b^{0}}} w^{0}_i \phi(\Xx_i)
+ \sum_{\Xx \in \XDomain_{0}^{j\neq b^{1}}} w^{1}_i \phi(\Xx_i),
\end{equation}
Intuitively, the highest and the second-highest skinning weights $w^{0}_i, w^{1}_i$ transition smoothly near the connection areas, and the dual-weighting mechanism takes advantage of this trait to bring gradual supervision to these areas.

\PAR{Latent Regularization.}
Similar to DeepSDF~\cite{park_deepsdf_2019}, we add a regularization term to the learnable shape codes to further constrain the learning problem:
\begin{equation}
    \Xloss_{\text{zs}}= \|\Xz_s\|_2^2.
\end{equation}
Our final loss is then defined as
\begin{equation}
\begin{split}
\Xloss_{\text{total}}= & \lambda_{\text{dual-m}}\Xloss_{\text{dual-m}} + \lambda_{\text{n}} \Xloss_{\text{n}} + \lambda_{\text{e}} \Xloss_{\text{e}} \\
 & + \lambda_{\text{nm}} \Xloss_{\text{nm}} + \lambda_{\text{osnm}} \Xloss_{\text{osnm}} + \lambda_{\text{zs}} \Xloss_{\text{zs}}.
\end{split}
\end{equation}
The values for the loss weights $\lambda$ can be found in the supplementary material.

\subsection{Implementation Details}
\label{ssec:implementation_details}
For each part decoder, we use 4 fully-connected layers with 64 neurons.
We initialize the network weights with the geometric initialization introduced by Atzmon~\etal~\cite{atzmon_sal_2020}.
Additionally we rely on weight normalization. 
We use the Adam optimizer~\cite{Kingma-Ba-ICLR-2015} with an initial learning rate of $0.0001$, and divide the learning rate by half every 500 epochs.


\section{Experiments}
\label{sec:experiments}
We evaluate our proposed representation on various tasks, including human body representation (Sec.~\ref{ssec:exp_representation}), shape interpolation (Sec.~\ref{ssec:exp_shape_interp}), model fitting (Sec.~\ref{ssec:exp_model_fitting}), human pose tracking and retargeting (Sec.~\ref{ssec:exp_pose_tracking}).
Furthermore, we also demonstrate our capability of fine-tuning with partial raw scan data (Sec.~\ref{ssec:exp_raw_scan_fine_tuning}).
At last, we perform ablation studies (Sec.~\ref{ssec:exp_ablation}) to analyze the efficacy of the proposed non-rigid geometric supervision and the latent optimization.

\subsection{Dataset}
We use the DFaust and MoVi subsets from the AMASS~\cite{mahmood_amass_2019} dataset to evaluate the model representation, and follow the split from LEAP~\cite{mihajlovic_leap_2021} for training and testing.
For the AMASS / DFaust subset, we use training sequences of 10 subjects and randomly keep 1 sequence per subject for testing.
For the AMASS / MoVi subset, we select every 10-th subject (which yields 9 unseen subjects) for testing and use the remaining 76 subjects for training in order to evaluate the generalization ability to unseen shapes.
To further evaluate our capability of dealing with non-watertight partial raw scan data, we fine-tune on
DFaust~\cite{dfaust} body scans and the CAPE~\cite{CAPE,clothcap} clothed human dataset, and randomly select two subjects from each dataset for evaluation.
Please refer to the supplementary material for more details.

\begin{figure}[t!]
    \centering
    \includegraphics[width=1.0\linewidth, trim={0 0 0 0}, clip]{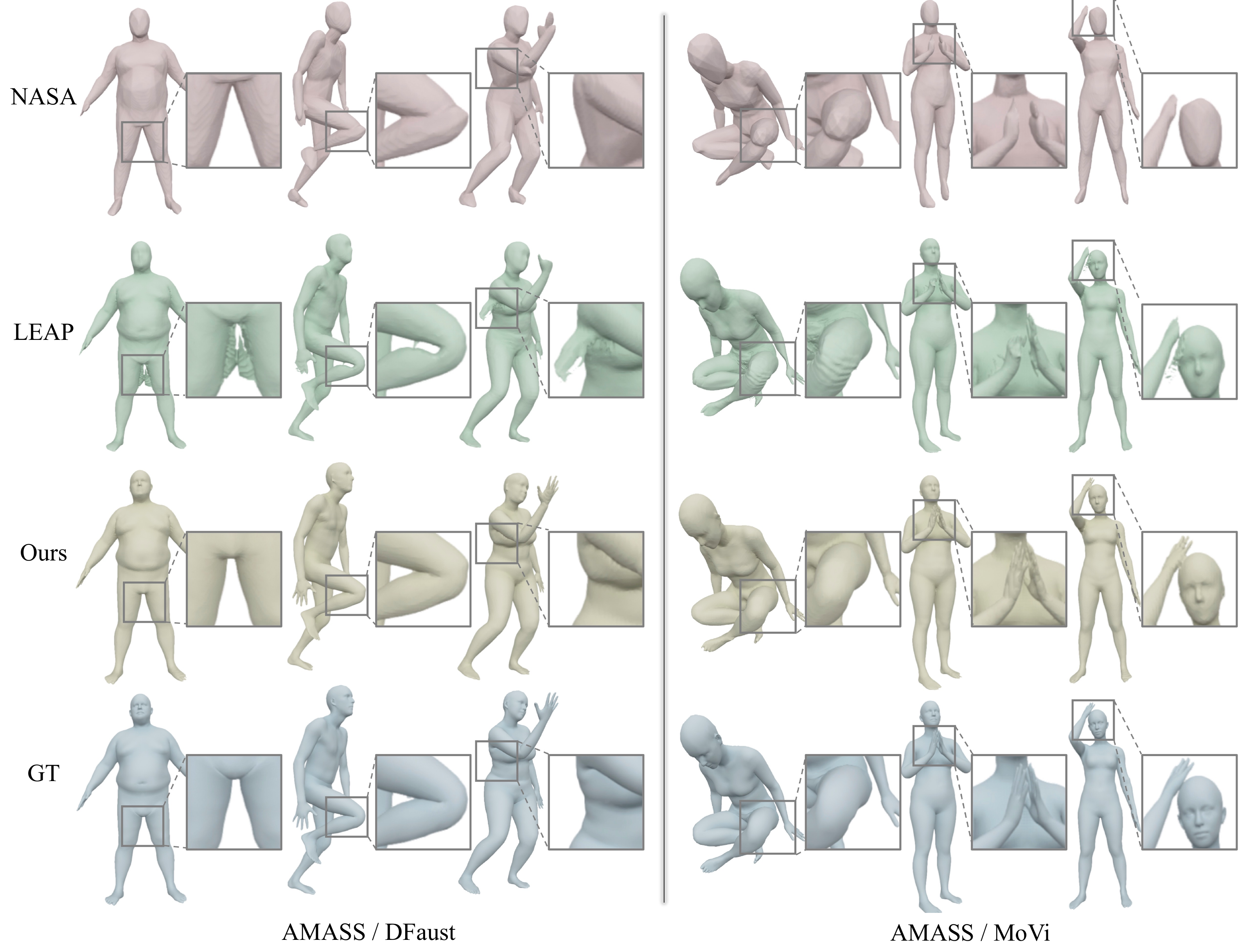}
    \caption{
    \textbf{Qualitative comparison.}
    We compare our representation qualitatively with NASA~\cite{deng_nasa_2020} and LEAP~\cite{mihajlovic_leap_2021} on the DFaust and MoVi subsets 
    of AMASS~\cite{mahmood_amass_2019}.
    Our method preserves more shape details than NASA. Compared with LEAP, our method mitigates artifacts, especially in cases where different body parts have great proximity (\eg hand and head on the far right).
    Best viewed digitally.
    }
    \label{fig:recon}
\end{figure}

\input{tables/tab_reconstruct}

\subsection{Human Body Representation}
\label{ssec:exp_representation}
To evaluate our representation on human bodies,
we compare our method with NASA~\cite{deng_nasa_2020} and LEAP~\cite{mihajlovic_leap_2021} \footnote{NPMs \cite{palafox_npms_2021} is a concurrent work without public code available  and SCANimate \cite{saito_scanimate_2021} aims to align raw scans of a particular person to learn an animatable avatar, so they are not adopted for comparison.} on AMASS / DFaust and AMASS / MoVi data.
We follow LEAP~\cite{mihajlovic_leap_2021} and use the same setup for the competitors.
For our method, it is worth noting that, when evaluating on the unseen subjects of AMASS / MoVi data, we only optimize the shape code for the first frame of each sequence and reconstruct the following frames with the fixed shape code.
For the AMASS / DFaust data, our method directly uses the learnt shape codes of the specific subjects from the training process.
For measuring the representation quality, we report the same metrics used in NASA~\cite{deng_nasa_2020}, including IoU (\%), Chamfer-L1 (m) and F-Score (\%).

As shown in Table~\ref{tab:body_recon}, our representation consistently outperforms NASA and LEAP on all metrics.
We also provide qualitative comparisons on the resulting meshes in Fig.~\ref{fig:recon}.
NASA pioneered neural implicit representations for articulated shapes like human bodies but only focused on representing one single subject at a time and therefore struggles to represent multiple subjects with the same network. This can be observed in the first row of Fig.~\ref{fig:recon} where the details for the face and hands are missing. 
In contrast, our method accurately represents the details of different subjects.

The recently proposed LEAP also shows promising results for the multi-subject case, but suffers from the limitation of the inverse LBS mechanism.
In cases where two body parts are close (\eg, sitting with cross-legs or clapping hands), their reconstruction results show concavities or some parts (\eg, fingers) are even missing. 
We believe that in such situations, the inverse LBS network fails to accurately map a query point to the correct canonical space, which inevitably distorts the occupancy representation as discussed in their paper.
In comparison, our method directly expresses the SDF values in the posed space rather than learning an inverse mapping to the canonical space and thus shows less susceptibility to such cases where body parts are close by, leading to fewer artifacts.
Note that LEAP requires much more input data (\eg posed SMPL mesh vertices) for each reconstruction while our representation only depends on a learnt shape code for each human identity.

\begin{figure}[t!]
    \centering
    \includegraphics[width=0.9\linewidth, trim={0 0 0 0}, clip]{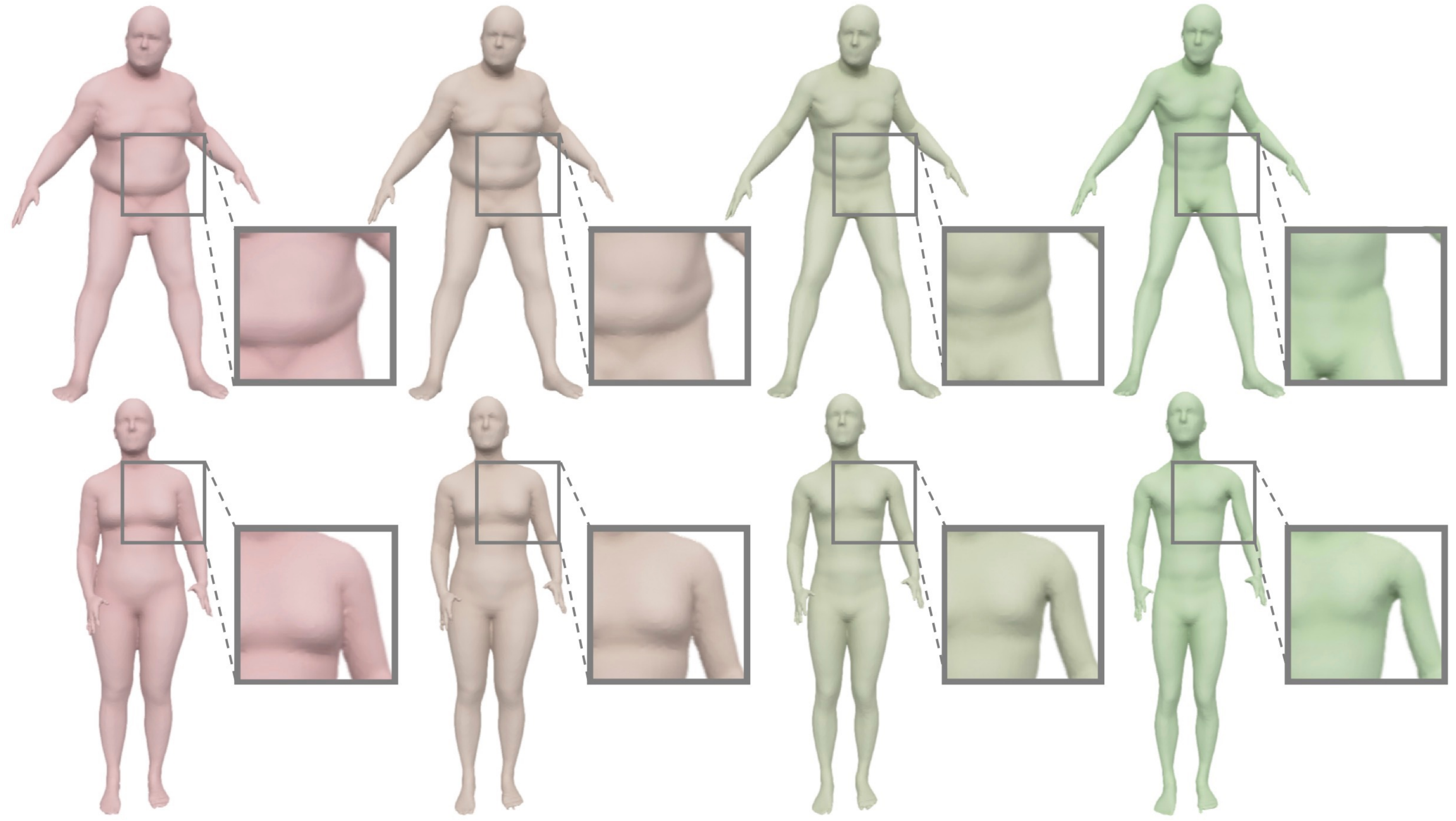}
    \caption{
    \textbf{Shape code interpolation.} We linearly interpolate the shape code of two examples from the AMASS / DFaust dataset (top row) and the AMASS / MoVi dataset (bottom row) respectively.
    }
    \label{fig:interpolation}
    \vspace{-0.7em}
\end{figure}

\subsection{Shape Interpolation}
\label{ssec:exp_shape_interp}

In order to demonstrate that the learnt shape representation is meaningful and continuous in latent space, we perform a shape code interpolation experiment.
Specifically, we choose two shape codes from Sec.~\ref{ssec:exp_representation} and interpolate linearly between them while having the pose fixed.
As visualized in Fig.~\ref{fig:interpolation}, the body shape transition is fairly smooth, which demonstrates that our shape code learns a meaningful and continuous representation of the human body.

\begin{figure}[t!]
    \centering
    \includegraphics[width=0.95\linewidth, trim={0 0 0 0}, clip]{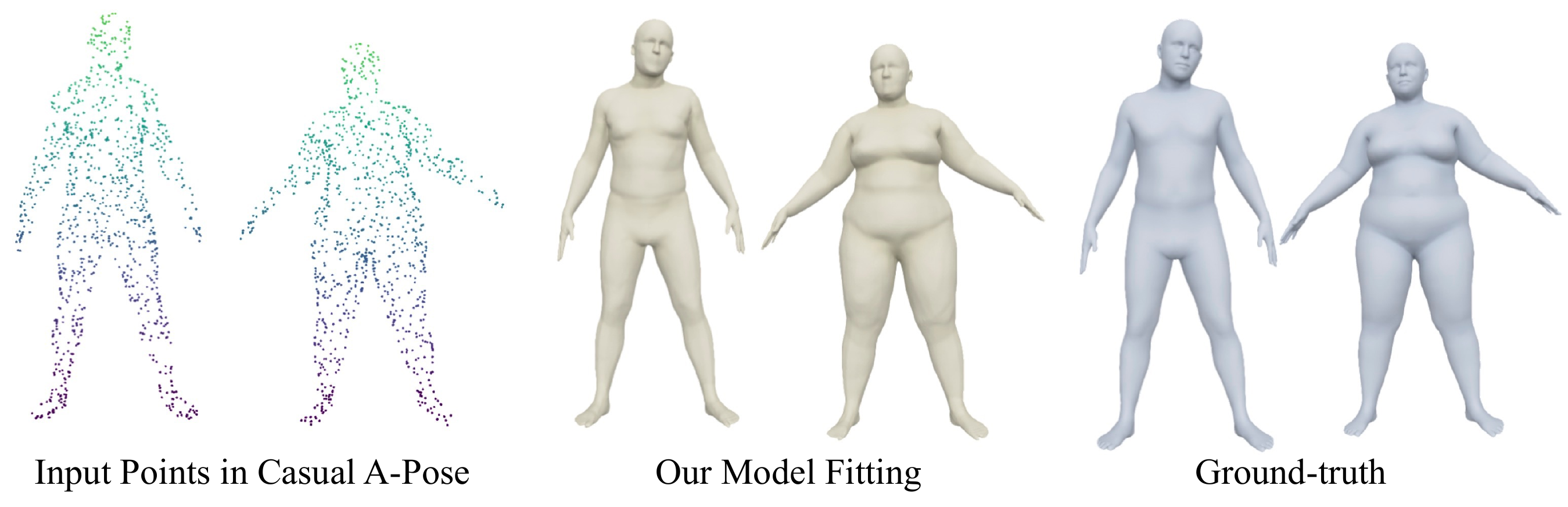}
    \caption{
    \textbf{Model fitting.} 
    We fit our representation to input points of human bodies in a casual A-pose by jointly optimizing the shape and pose. 1000 points were used as input.\looseness=-1
    }
    \label{fig:model_fitting}
\end{figure}

\subsection{Model Fitting}
\label{ssec:exp_model_fitting}

In this experiment, we demonstrate how our network can also be applied to model fitting tasks, \eg, fitting our representation to a given point cloud of human bodies in the casual A-pose, which is commonly used when capturing human bodies.
For the experimental setup, previously unseen body shapes from the SURREAL~\cite{surreal} dataset are randomly selected, combined with casual A-poses from the AMASS / DFaust dataset~\cite{mahmood_amass_2019} and uniformly sampled resulting in 1000 points on the surface.
For the fitting, we first initialize the shape code with a random normal distribution and the pose code with the standard A-pose encoded by VPoser~\cite{pavlakos_expressive_2019}.
We then jointly optimize both, the shape and pose code, by minimizing a single overall manifold loss.
As shown in Fig.~\ref{fig:model_fitting}, the reconstructed human bodies are intact and close to the ground-truth, which demonstrates the potential usefulness of our proposed representation for real-world applications.
Once the disentangled shape and pose codes are recovered, we can easily swap the shape to other identities by exchanging the shape code and animate the human body with novel poses as shown in Fig.~\ref{fig:teaser}.

\begin{figure}[t!]
    \centering
    \includegraphics[width=1.0\linewidth, trim={0 0 0 0}, clip]{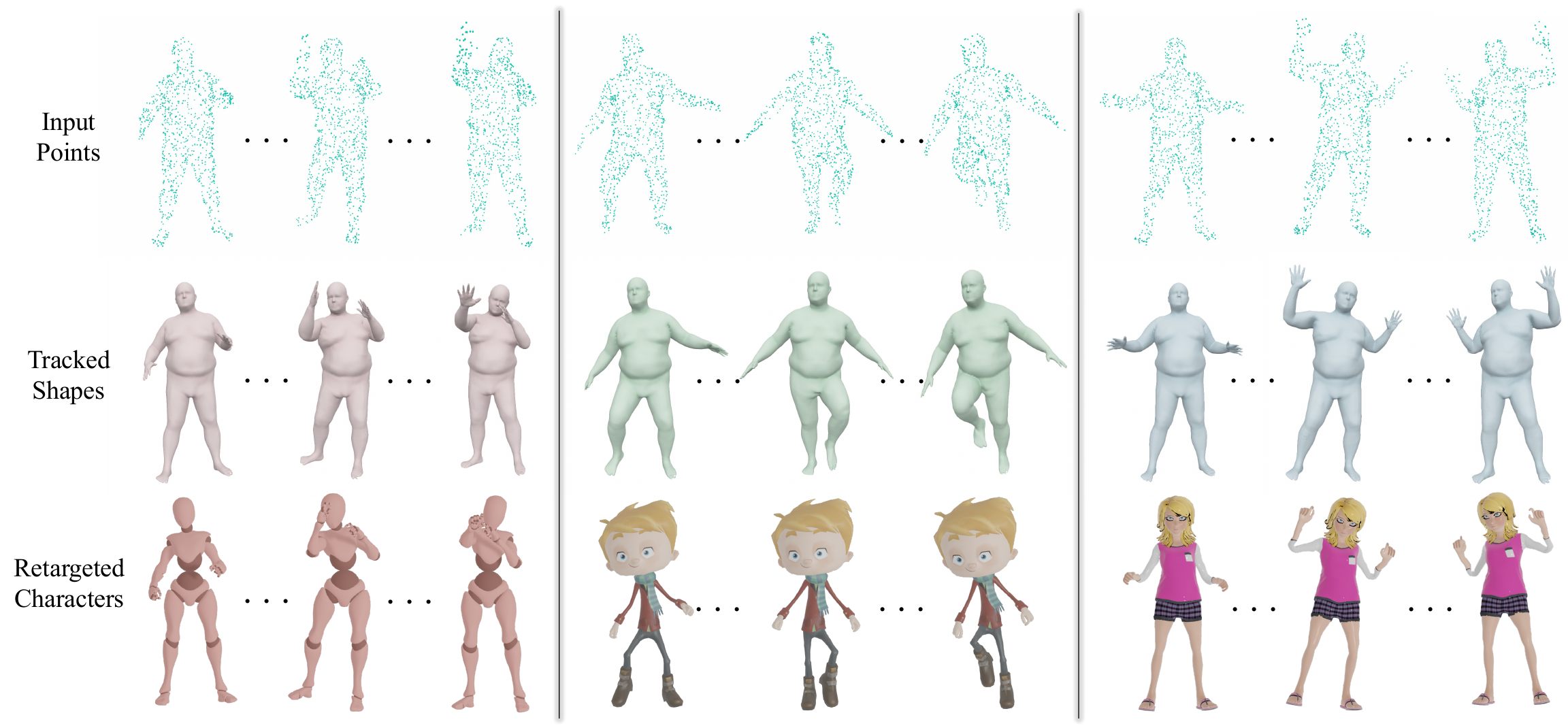}
    \caption{
    \textbf{Qualitative pose tracking results.}
    We show the input sparse points (first row), tracked human bodies with our reconstruction (second row), and the retargeted characters with the tracked poses (third row).
    }
    \label{fig:tracking}
\end{figure}

\input{tables/tab_pose_tracking}

\subsection{Human Pose Tracking}
\label{ssec:exp_pose_tracking}

As our representation is fully differentiable, it is readily available for human pose tracking on point clouds.
We choose three sequences from the AMASS / DFaust subset, and provide a comparison between our method and NASA with the pre-trained model from Sec.~\ref{ssec:exp_representation}.
For our method, we fix the learned shape code from Sec.~\ref{ssec:exp_representation} and optimize only the pose code directly within the latent space.
It is worth noting that we do not add any temporal or prior constraints like NASA does, but instead only use a single overall manifold loss and initialize the pose code with the optimized code from the previously tracked frame.
We evaluate the tracking performance on different point sampling densities, \ie, from 4000 to 500, and also follow NASA's setup by adding normal noise with $\sigma=0.005$ to the points.
The results are shown in Table~\ref{tab:pose_tracking},
where we report the IoU (\%) and MPJPE (m) (mean per joint position error~\cite{ionescu2013human3}) metrics.
Compared to NASA, we consistently achieve the best IoU and MPJPE in all settings.
We also visualize our tracked shapes and show motion retargeting to cartoon characters as an application in Fig.~\ref{fig:tracking}, demonstrating the meaningfulness and usefulness of the tracked poses. 
Please refer to our supplementary material for more results.

\subsection{Evaluation on Partial Raw Scan}
\label{ssec:exp_raw_scan_fine_tuning}
\begin{figure}[t!]
    \centering
    \includegraphics[width=1.0\linewidth, trim={0 0 0 0}, clip]{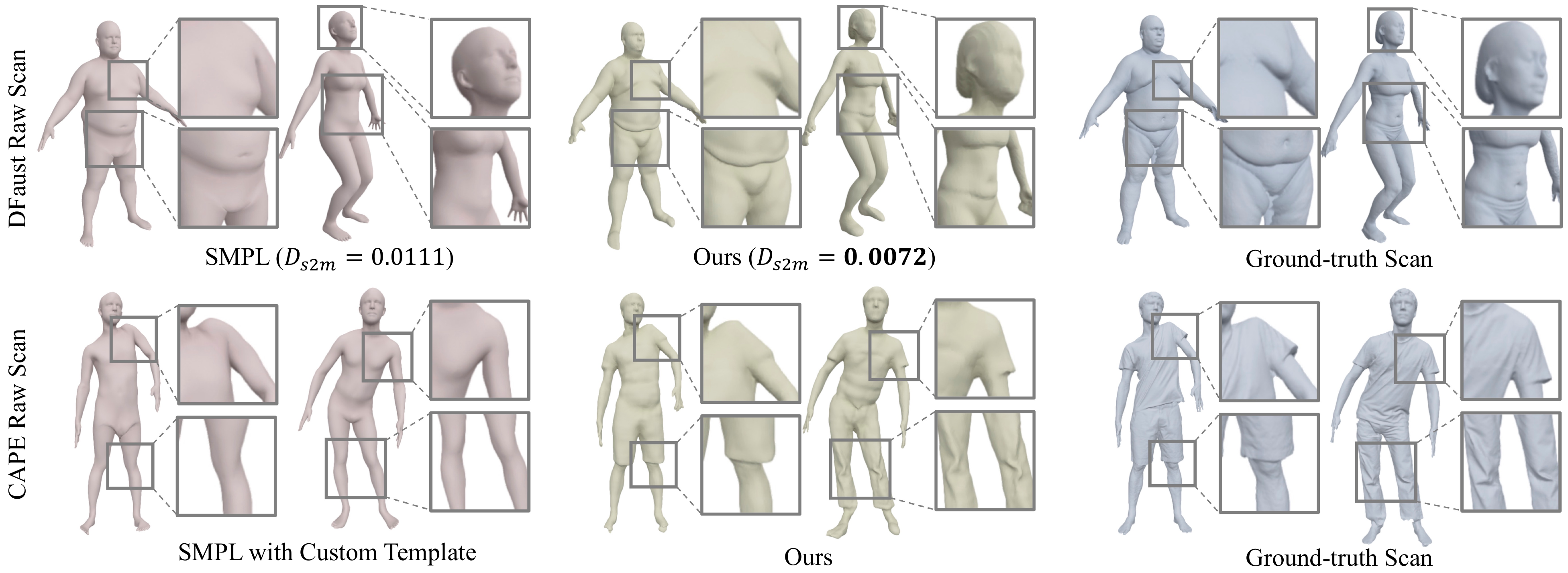}
    \caption{
    \textbf{Fine-tuning on raw scans.} Our non-rigid geometric supervision formulation allows us to fine-tune our model on raw partial scans. We are able to better preserve details than SMPL~\cite{loper_smpl_2015} on raw DFaust~\cite{dfaust} scans and even can model clothed human from the CAPE~\cite{CAPE,clothcap} dataset.
    }
    \label{fig:fine_tune}
    \vspace{0.3em}
\end{figure}

To further inspect the capability of our representation to handle partial raw scan data,
we fine-tune the pre-trained model (from Sec.~\ref{ssec:exp_representation}) on DFaust~\cite{dfaust} body scans and the CAPE~\cite{CAPE,clothcap} clothed human dataset.
Thanks to our non-rigid supervision, we are able to learn the shape naturally from these partial scans.
As shown in Fig.~\ref{fig:fine_tune}, our method better represents details and captures more realistic shapes of human bodies than SMPL on DFaust body scans (\eg, our head shape is closer to the ground-truth), and also achieves better one-sided Chamfer distance $D_{s2m}$ (\ie from scan to the resulting mesh following~\cite{gropp_implicit_2020}) than SMPL.
More importantly, our representation is able to successfully represent clothed human bodies on the CAPE dataset, even though the accessible training data of CAPE is limited, \ie, only 2 or 3 sequences are available for a few of released subjects.
Please refer to the supplementary video for a vivid animation of these results. 

\subsection{Ablation Study}
\label{ssec:exp_ablation}
\vspace{0.3em}

\input{tables/tab_ablation_main}

\begin{figure}[t!]
    \centering
    \includegraphics[width=0.9\linewidth, trim={0 0 0 0}, clip]{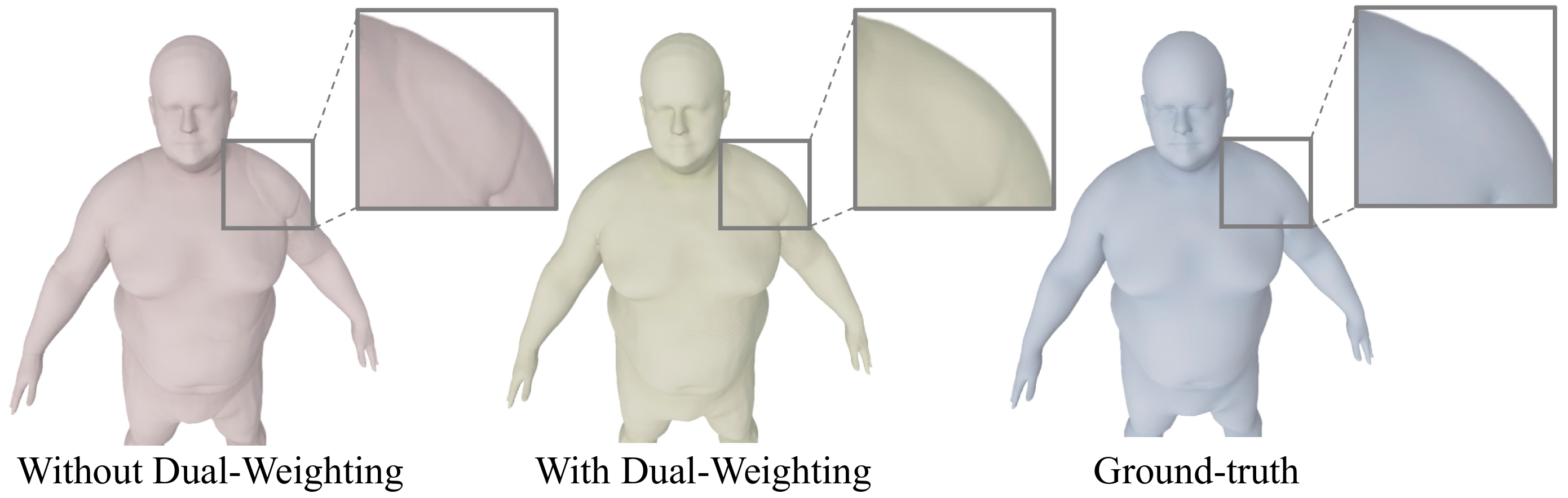}
    \caption{
    \textbf{Dual-weighting.} The dual-weighting strategy significantly reduces visual boundary artifacts at body part connection areas.
    }
    \label{fig:dual_weighting}
    \vspace{0.3em}
\end{figure}

\PAR{Dual-weighting.}
We analyze the efficacy of the dual-weighting mechanism by ablating it during the training process.
Even though the proposed strategy only slightly improves the overall metric scores (Table~\ref{tab:ablation}), it leads to a significant visual improvement by eliminating boundary artifacts near part connection areas (Fig.~\ref{fig:dual_weighting}).

\PAR{One-sided Loss.}
To study the impact of the one-sided loss, we remove it from the supervision during training.
As demonstrated in Table~\ref{tab:ablation}, the network does not converge to form a human body due to the lack of polarized SDF supervision as explained in Sec.~\ref{ssec:non_rigid_geometric_supervision}, which highlights the importance of this loss for our non-rigid geometric supervision.

\input{tables/tab_ablation_pose_code}
\vspace{0.3em}

\PAR{VJointer on Shape Code Optimization.}
Since our experiment on the AMASS / MoVi subset of Sec.~\ref{ssec:exp_representation} leverages the network's ability of shape code optimization, we analyze the effectiveness of VJointer by removing it and instead optimizing the $B\times3$ joint locations directly.
As shown in Table~\ref{tab:ablation}, the proposed VJointer significantly improves the reconstruction quality by a large margin.

\PAR{Latent Pose Optimization.}
We inspect the effectiveness of latent pose optimization on the human pose tracking task of Sec.~\ref{ssec:exp_pose_tracking} by replacing it with a direct optimization of the bone transformations, \ie, optimizing the rotation with a continuous representation~\cite{zhou2019continuity}. 
The tracking results are reported in Table~\ref{tab:ablation_pose_code}.
Introducing pose optimization on the latent space significantly enhances the tracking robustness.

\section{Conclusion}

We introduced LatentHuman, a novel pose-shape-disentangled representation for the human body.
We exploit both pose and shape priors by using the kinematic model for guiding the network to disentangle the shape and pose spaces.
Our non-rigid geometric supervision also allows to fine-tune LatentHuman on non-watertight raw scans.
Experiments demonstrate that LatentHuman outperforms existing methods for 3D human body representation and can be applied to various tasks like 3D model fitting, shape interpolation, and pose tracking.
Although LatentHuman can model the clothed human body, it cannot handle the soft cloth dynamics and we leave this as a future work.

\noindent\textbf{Acknowledgments:}
This work was partially supported by the NSFC (No.~62102356), Zhejiang Lab (2021PE0AC01) and Innosuisse (Grant No. 34475.1 IP-ICT).
 
\clearpage
{\small
\bibliographystyle{ieee_fullname}
\bibliography{deephuman}
}

\end{document}

%% file: tables/tab_contributions.tex
\begin{table}[!t]
\centering
\ra{1.3}
\newcommand{\cmk}{\checkmark}
\newcommand{\inpSC}{\ensuremath{\mathbf{z}_{s}}} 
\newcommand{\inpSP}{\ensuremath{\beta}} 
\newcommand{\inpPC}{\ensuremath{\mathbf{z}_{p}}} 
\newcommand{\inpPP}{\ensuremath{\theta}} 
\newcommand{\inpPV}{\ensuremath{\hat{V}}} 
\resizebox{\columnwidth}{!}{
    \begin{tabular}{@{}lccHcHccccc@{}}
        \toprule
               &     & Test   &       & Shape  & & \multicolumn{2}{c}{Pose} & Non-& Raw   & Mutli. \\
        Method & Out & Input  & Arch. & Optim. & & Optim. & Contr.          & WaTi.& Scan  & Shape \\
        \midrule
        NASA~\cite{deng_nasa_2020}            & Occ. & \inpPP{}                    & AD & & & \cmk & \cmk & & & \\
        LEAP~\cite{mihajlovic_leap_2021}      & Occ. & \inpSP{},\inpPP{},\inpPV    & & & & & \cmk & & \cmk & \cmk \\ 
        SCANimate~\cite{saito_scanimate_2021} & SDF  & \inpPP{}                    & & & & & \cmk & \cmk & \cmk & \\ 
        NPMs~\cite{palafox_npms_2021}         & SDF  & \inpSC{},\inpPC{}           & & \cmk & & \cmk & & & \cmk & \cmk \\ 
        Ours                                  & SDF  & \inpSC{}, \inpPP{} or \inpPC{} & Hier. AD & \cmk & \cmk & \cmk & \cmk & \cmk & \cmk & \cmk \\
        \bottomrule
    \end{tabular}
}
\vspace{-3pt}
\caption{\textbf{Overview of implicit representations for human bodies.} 
We denote whether shape and pose parameters are optimizable (\eg, for pose-tracking) and/or controllable (\eg, for animation) 
and compare whether the model can be trained on non-watertight data, generalize to raw scans and be reused for multiple subjects without re-training.
(\inpSP{},\inpPP{}): SMPL-based shape/pose parameters, (\inpSC{},\inpPC{}): Learnt shape/pose latent codes, \inpPV{}: Posed SMPL vertices.
}
\label{tab:contribution}
\end{table}

%% file: tables/tab_reconstruct.tex
\begin{table}[tb]
\centering
\resizebox{1.0\linewidth}{!}{
\begin{tabular}{lcccccc}
\toprule
\multicolumn{1}{c}{\multirow{2}{*}{Methods}} & \multicolumn{3}{c}{AMASS / DFaust} & \multicolumn{3}{c}{AMASS / MoVi} \\ \cmidrule(lr){2-4} \cmidrule(lr){5-7}  
\multicolumn{1}{c}{} & \multicolumn{1}{l}{IoU $\uparrow$} & \multicolumn{1}{l}{Chamfer $\downarrow$} & \multicolumn{1}{l}{F-Score $\uparrow$} & \multicolumn{1}{l}{IoU $\uparrow$} & \multicolumn{1}{l}{Chamfer $\downarrow$} & \multicolumn{1}{l}{F-Score $\uparrow$} \\ \midrule
NASA~\cite{deng_nasa_2020} & 87.67 & 0.00719 & 80.95 & 84.08 & 0.00885 & 73.36 \\
LEAP~\cite{mihajlovic_leap_2021} & 96.09 & 0.00333 & 98.24 & 94.72 & 0.00352 & 98.23 \\
Ours & \textbf{96.45} & \textbf{0.00304} & \textbf{99.06} & \textbf{95.76} & \textbf{0.00314} & \textbf{98.90} \\
\bottomrule
\end{tabular}
}
\caption{
\textbf{Quantitative reconstruction results.} 
Our representation outperforms SoTA approaches~\cite{deng_nasa_2020,mihajlovic_leap_2021} on all metrics.
}
\label{tab:body_recon}
\end{table}

%% file: tables/tab_pose_tracking.tex
\begin{table}[tb]
\centering
\resizebox{1.0\linewidth}{!}{
\begin{tabular}{lcccc}
\toprule
\multicolumn{1}{c}{\textbf{\# Sampled Points}} & \textbf{4000} & \textbf{2000} & \textbf{1000} & \textbf{500} \\
\midrule
\multicolumn{5}{c}{IoU (\%) $\uparrow$ / MPJPE (m) $\downarrow$} \\
\midrule
NASA~\cite{deng_nasa_2020} & 92.99 / 0.0150 & 92.86 / 0.0156 & 92.59 / 0.0162 & 92.12 / 0.0178 \\
Ours & \textbf{95.88} / \textbf{0.0049} & \textbf{95.77} / \textbf{0.0051} & \textbf{95.53} / \textbf{0.0057} & \textbf{95.20} / \textbf{0.0066} \\
\bottomrule
\end{tabular}
}
\caption{
\textbf{Quantitative pose tracking results.} We report IoU and mean per-joint position error (MPJPE~\cite{ionescu2013human3}) on four different point densities.
}
\label{tab:pose_tracking}
\end{table}

%% file: tables/tab_ablation_main.tex
\begin{table}[tb]
\centering
\resizebox{1.0\linewidth}{!}{
\begin{tabular}{lcccccc}
\toprule
\multicolumn{1}{c}{\multirow{2}{*}{Methods}} & \multicolumn{3}{c}{AMASS / DFaust} & \multicolumn{3}{c}{AMASS / MoVi} \\ \cmidrule(lr){2-4} \cmidrule(lr){5-7}  
\multicolumn{1}{c}{} & \multicolumn{1}{l}{IoU $\uparrow$} & \multicolumn{1}{l}{Chamfer $\downarrow$} & \multicolumn{1}{l}{F-Score $\uparrow$} & \multicolumn{1}{l}{IoU $\uparrow$} & \multicolumn{1}{l}{Chamfer $\downarrow$} & \multicolumn{1}{l}{F-Score $\uparrow$} \\ \midrule
w/o Dual-Weighting & 96.31 & \textbf{0.00310} & 98.96 & 94.29 & 0.00532 & 96.34 \\
w/o One-Sided Loss & $\times$ & $\times$ & $\times$ & $\times$ & $\times$ & $\times$ \\
w/o VJoints & / & / & / & 89.44 & 0.04796 & 91.45 \\
Full Model & \textbf{96.45} & \textbf{0.00304} & \textbf{99.06} & \textbf{95.76} & \textbf{0.00314} & \textbf{98.90} \\
\bottomrule
\end{tabular}
}
\caption{
\textbf{Ablation study.} We ablate our dual-weighting strategy, the one-sided manifold loss and our proposed VJointer module. All proposed components are necessary to achieve the best performance.
}
\label{tab:ablation}
\end{table}

%% file: tables/tab_ablation_pose_code.tex
\begin{table}[tb]
\centering
\resizebox{1.0\linewidth}{!}{
\begin{tabular}{lcccc}
\toprule
\multicolumn{1}{c}{\textbf{\# Sampled Points}} & \textbf{4000} & \textbf{2000} & \textbf{1000} & \textbf{500} \\
\midrule
\multicolumn{5}{c}{IoU (\%) $\uparrow$ / MPJPE (m) $\downarrow$} \\
\midrule
Ours (w/o Latent) & 95.85 / 0.1520 & 94.17 / 0.1762 & 93.28 / 0.2299 & $\times$ \\
Ours (w Latent) & \textbf{95.88} / \textbf{0.0049} & \textbf{95.77} / \textbf{0.0051} & \textbf{95.53} / \textbf{0.0057} & \textbf{95.20} / \textbf{0.0066} \\
\bottomrule
\end{tabular}
}
\caption{
\textbf{Ablation on latent pose code optimization.} We perform pose tracking with and without latent pose code optimization, \ie instead directly optimizing bone transformations. Optimizing the pose latent space leads to significant improvements.
}
\label{tab:ablation_pose_code}
\end{table}